\documentstyle[cgloss4e,fullname]{article}
\makeatletter
 \def\submitted#1{\setbox\@tempboxa\vbox{\normalsize \tt \raggedright
    #1 \\ \hbox{}}
    \vspace{-2.5 cm} \usebox\@tempboxa \\
    \vspace{-\ht\@tempboxa} \vspace{2.5 cm}}
\makeatother

\author{Barbara Di Eugenio\\
Learning Research and Development Center \\
University of Pittsburgh\\
Pittsburgh, PA, 15260 USA\\ {\tt dieugeni@cs.pitt.edu}}

\title{\submitted{To appear in "Centering in Discourse", Oxford
University Press, 1997}Centering in Italian}

\newcommand{\bdes}{\begin{description}}
\newcommand{\edes}{\end{description}}
\newcommand{\bite}{\begin{itemize}}
\newcommand{\eite}{\end{itemize}}
\newcommand{\benum}{\begin{enumerate}}
\newcommand{\eenum}{\end{enumerate}}
\newcommand{\bverb}{\begin{verbatim}}
\newcommand{\everb}{\end{verbatim}}

\newcommand{\m}[1]{{\em #1\/}}

\newcounter{thesubeg}
\newcounter{thesubegfoo}
\newenvironment{egs}[1]{\refstepcounter{equation}\label{#1}\samepage\begin{list}
{(\arabic{equation}\alph{thesubeg})}{\usecounter{thesubeg}}}{\end{list}}
\newenvironment{eg}[1]{\refstepcounter{equation}\label{#1}\samepage\begin{list}
{(\arabic{equation})}{\usecounter{thesubeg}}}{\end{list}}
\newcommand{\refegs}[2]{\setcounter{thesubegfoo}{#2}(\ref{#1}\alph{thesubegfoo})}
\newcommand{\refeg}[1]{(\ref{#1})}
\newcommand{\noparrefegs}[2]{\setcounter{thesubegfoo}{#2}\ref{#1}\alph{thesubegfoo}}

\newcommand{\centc}{{\sc continue\/}}
\newcommand{\centcr}{{\sc ret-cont\/}}
\newcommand{\centr}{{\sc retain\/}}
\newcommand{\cents}{{\sc shift\/}}
\newcommand{\centss}{{\sc smooth-shift\/}}
\newcommand{\centrs}{{\sc rough-shift\/}}

\newcommand{\tref}[1]{Table~\ref{#1}}

\date{}

\begin{document}

\maketitle
\bibliographystyle{fullname}

\noindent
{\bf Abstract.} This chapter explores the correlation between
centering and different forms of pronominal reference in Italian, in
particular zeros and overt pronouns in subject position. In previous
work \cite{coling90}, I proposed that such alternation could be
explained in terms of centering transitions. In this chapter, I verify
those hypotheses by means of a small corpus of naturally occurring
data. In the process, I extend my previous analysis in several ways,
for example by taking possessives into account; I also provide a more
detailed analysis of {\sc continue}: more specifically, I show that
pronouns are used in a markedly different way in {\sc continue}'s
preceded by another {\sc continue} or by a \cents, and in {\sc
continue}'s preceded by a \centr.

\section{Introduction}

Italian is a \m{pro-drop} language, in that the subject of a clause
need not be overt. Thus, an Italian speaker has a variety of choices
in realizing a subject, including using a null subject or an overt
pronoun.  In previous work \cite{coling90}, I proposed that the
alternation of null and overt pronominal subjects could be explained
in terms of centering transitions. However, the hypotheses I put
forward in my earlier work were supported only by a few constructed
examples.  In this chapter, I report on a corpus study that I
conducted in order to find more solid evidence for those
hypotheses. Analyzing real data had the added benefit of bringing me
to address issues still problematic for centering, such as how
possessives and subordinates affect the ordering on the Cf list, and
to provide a more detailed analysis of {\sc continue}'s.

The version of centering that I use is basically the one described in
the overview by Walker, Prince and  Joshi (this volume). However, the
ordering for the Cf list I use is modified with respect to
the usual one postulated for   Western languages:
\begin{eg}{cf-ranking}
\item \ {\sc subject $>$ object2 $>$ object $>$ others}
\end{eg}
Various researchers \cite{kame,lyn3} had pointed out that in languages
such as Japanese, \m{empathy} and \m{topic} marking affect Cf ordering.  Turan
in \shortcite{turan95} argues that the notion of {\em empathy} is
relevant to Western languages as well, because of non-agentive
psychological verbs, such as \m{interest, seem}; perception verbs,
such as \m{feel, appear}; and in general, expressions that refer to a
character's point of view, such as \m{The thought crossed her
mind}.\footnote{For a thorough treatment of subjective expressions and
tracking characters' points of view, see \cite{wiebe94}.} Turan points
out that, with such expressions, it is the experiencer, which is often
in object position, rather than the grammatical subject, that should
be ranked higher. Moreover, Turan points out that in her Turkish
corpus quantified indefinite subjects (qis) and arbitrary plural pro's
(pro$_{arb}$) rank very low. Therefore, the Cf ranking in
\refeg{cf-ranking} should be amended as follows
\cite{turan95}:\footnote{The first researcher to propose that {\em empathy}
should precede {\em subject} in the Cf ranking was \cite{kame}.}
\begin{eg}{cf-ranking2}
\item \ {\sc {\bf empathy} $>$  subject $>$ object2 $>$ object $>$ others
$>$ {\bf qis, pro$_{arb}$}}
\end{eg}
In this chapter, I will adopt \refeg{cf-ranking2}.

Another difference between the standard notion of centering and the
way I applied it is that I don't explicitly take discourse segment boundaries
into account. According to \cite{gjw95}, centering is a local
mechanism that applies within a single specific discourse segment, but
not  across segment boundaries. However, segmenting discourse
is an active research area in itself, and there are no texts with agreed upon
discourse structures. It seems that, when analyzing naturally occurring
text, two approaches are possible:
\bite
\item The first is to postulate a segment structure for the text of
interest, for example exploiting paragraph boundaries, cue words etc
\cite{walker89}.
\item The second is to disregard segment boundaries and apply
centering between every two adjacent utterances.  It is in fact possible that
the absence of a centering transition between two utterances
indicates a segment boundary --- see Passonneau (this volume), Walker (this
volume). This is the  approach I adopted, as my interest is in using centering
as an explanatory tool for the distribution of  pronominal forms.
Notice that  the cues that \cite{walker89} uses to provide a
discourse structure prior to applying centering include whether the
first sentence of a new paragraph 
\begin{quote}
has a pronoun in subject position or a pronoun where none of the
preceding sentence-internal noun phrases match its syntactic features.
\end{quote}
In these cases, Walker doesn't consider the  paragraph as
constituting a  discourse segment separate from the preceding one.
\eite

Finally, note that in this paper, as in my previous work,
I take the same position advocated in \cite{lyn3},
\begin{quote}
that the interpretation of zeros is an inferential
process, but that syntactic information provides constraints on this
inferential process.
\end{quote} 
I will suggest that it is the syntactic context up and including the
verbal complex that affects the interpretation of null
subjects.\footnote{I'm  using here the term  {\em inferential processing},
and later terms such as \m{strategies}, in their intuitive sense.
Susan Brennan (p.c.) brought to my attention the difference between
\m{strategic} or \m{inferential} processing and \m{automatic}
processing, and the fact that syntactic cues of the kind I discuss in
this paper affect the latter, not the former.}

The chapter is organized as follows. In Sec.~\ref{pro-sec} I discuss
the Italian pronominal system and the hypotheses from my previous
work.  Sec.~\ref{corpus} describes the corpus I used and details the
distributions of various referring expressions in subject position.
In Sec.~\ref{cent-trans} I first discuss assumptions and extensions I
had to make in order to apply centering to naturally occurring text;
and then report on the correlations between pronouns and centering
transitions. Sec.~\ref{discuss} analyzes such correlations: in
particular, I refine the notion of {\sc continue} in order to account
for a non negligible number of occurrences of   strong pronouns. Finally,
Sec.~\ref{conclude} presents conclusions and future work.

\section{Italian pronouns and Centering}

\label{pro-sec}

\subsection{The Italian pronominal system}

In Italian, there are two pronominal systems, characterized by a
different syntactic distribution: weak pronouns, that must always be
cliticized to the verb ({\sf la, lo, li, le, gli} - respectively her,
accusative; him, accusative; them, masculine accusative; them,
feminine accusative or her, dative; him, dative), and strong pronouns
({\sf lui, lei, loro} - respectively he or him; she or her; they or
them).\footnote{Traditionally, {\sf lui, lei, loro} were the oblique
forms of the strong system, with the nominative forms being
respectively {\sf egli, ella, essi/e}: \label{egli} however, in
current Italian the latter forms are only rarely used as the oblique
forms have replaced them in subject position --- among the 33
instances of strong pronouns in \tref{numbers3}, there are only three
occurrences of {\sf egli}, and all three of them occur in
\cite{dick}.} The null subject is considered part of the system of
weak pronouns.

In Italian there is no neuter gender: nouns referring to inanimate
objects are masculine or feminine.  The weak pronouns used in this
case are those of the corresponding gender.  However, strong pronouns
can't refer to inanimate objects, so that paraphrases or deictics are
used: a strong pronoun for inanimate objects does exist --- {\sf esso / essi}
for masculine, singular and plural, {\sf essa / esse} for feminine,
singular and plural --- but it is not much used in
current Italian: there is only one instance of {\sf esse} in my corpus,
in \cite{dick}.\footnote{This is the same article in which also the 
three occurrences of {\sf egli} appear, see fn.~\ref{egli}.}  

Weak and strong
pronouns are often in complementary distribution, as 
strong pronouns have to be used in prepositional phrases, e.g. {\sf
per lui}, \m{for him}, as in Ex.~\refegs{per-lui}{2}. However, this
syntactic alternation doesn't apply in subject position: the choice of 
null versus strong pronoun depends on pragmatic factors, and can be
accounted for in terms of centering transitions.

\begin{egs}{per-lui}
\item 
\gll  {\bf Maria$_i$}  \'{e} andata in vacanza con {\bf suo} {\bf padre$_j$}:\\
{\bf Maria$_i$}  is gone on holiday with {\bf her} {\bf father$_j$}:\\
\item 
\gll $\phi$ \'{e} stato un vero piacere per {\bf *lo$_j$/lui$_j$}.\\
(it) is been a real treat for {\bf him$_j$}.\\
\end{egs}

\subsection{Previous results}

In \cite{coling90}, I proposed that 
\begin{egs}{strategies}
\item Typically, a null subject signals  a \centc, and a strong pronoun
a \centr\/ or a \cents.
\item A null subject can be felicitously used in cases of \centr\/
or \cents\/ if $U_{i}$ provides syntactic features that force the null
subject to refer to a referent different from Cb$(U_{i-1})$. Moreover,
it is the syntactic context up to and including the verbal form(s)
carrying tense and / or agreement that makes the reference felicitous
or not.
\end{egs}

These claims stemmed from (constructed) examples such as the
following, where I use referents of different gender --- {\sf Maria},
female proper name; {\sf Giovanni, Giorgio}, male proper names --- to
show how gender and morphological markings come into play when
resolving reference.  These examples are not ambiguous in English,
given that a null subject is not an option available to a speaker.
Boldface is just meant to highlight referential expressions, not to
indicate stress; pronouns in parentheses in English correspond to
zeros\footnote{I will occasionally use the term \m{zero}: the speaker
should keep in mind that Italian allows to drop only subjects, at
least as a general rule.} in Italian; also remember that {\sf lui,
gli, lo} are masculine, and {\sf lei, le, la} feminine.

\begin{egs}{mare}
\item  
\gll {\bf Maria}$_i$ voleva andare al mare.\\
{\bf Maria}$_i$ wanted {to go} {to the} seaside.\\
\item 
\gll {$\phi_i$} Telefono' a  {\bf Giovanni}$_j$.\\
{\bf (She$_i$)} phoned to  {\bf Giovanni}$_j$.\\
\item  
\renewcommand{\theenumi}{\roman{enumi}}
\benum
\item  \label{mare1.1}
\gll {$\phi_i$} Si arrabbio' perche' {$\phi_i$} non {\bf lo}$_j$ trovo' a casa. \\
{\bf (She$_i$)} self {got angry} because {\bf (she$_i$)} not  {\bf him}$_j$  found at home.\\
\item   \label{mare1.2}
\gll $\phi_{i/?j}$ Si arrabbio' perche'  $\phi_j$ stava dormendo.\\
{\bf (She$_i$)}/{\bf (?He$_j$)} self {got angry}  because {\bf (he$_j$)} was sleeping.\\
\item \label{mare1.3}
\gll {\bf Lui}$_j$ si arrabbio' perche' $\phi_j$ stava dormendo.\\
{\bf He}$_j$  self {got angry} because {\bf (he$_j$)} was sleeping.\\
\item \label{mare1.4}
\gll $\phi_j$ Si e' arrabbiato$_{masc}$  perche'  $\phi_j$ stava dormendo.\\
{\bf (He$_j$)} self is {become angry$_{masc}$} because {\bf (he$_j$)} was sleeping.\\
\eenum
\end{egs}

\noindent
Consider the four \refegs{mare}{3} variations; notice that {\sf Maria}
is both Cb\refegs{mare}{2} and Cp\refegs{mare}{2}:
\bdes
\item[\refegs{mare}{3}.i] The null subject refers to {\sf Maria},
which is then both Cb(\noparrefegs{mare}{3}.i) and Cp(\noparrefegs{mare}{3}.i).
\refegs{mare}{3}.i thus
realizes a  {\sc continue}. 
\item[\refegs{mare}{3}.ii] The most natural interpretation is that 
the null subject in the main clause refers to {\sf Maria} --- the null
subject in the subordinate clause is forced to refer to {\sf Giovanni}
on pragmatic grounds.\footnote{In \cite{coling90} I was not
addressing the issue of interpreting  null subjects in
subordinate  clauses.} 
For this same pragmatic reason, 
the null subject in the main clause may be interpreted as referring
to {\sf Giovanni}, but the discourse sounds less coherent. 
\item[\refegs{mare}{3}.iii] As {\sf Giovanni} was neither Cb\refegs{mare}{2},  
nor Cp\refegs{mare}{2}, S performs  a felicitous 
\centss\/ by referring to {\sf Giovanni} with a strong pronoun.
\item[\refegs{mare}{3}.iv] Contrast this utterance with \refegs{mare}{3}.ii.
They should have the same effect on the hearer, namely, the null
subject should be interpreted as referring to {\sf Maria}: instead in
\refegs{mare}{3}.iv it is felicitously interpreted as referring to
{\sf Giovanni}.  This is due to the fact that in \refegs{mare}{3}.iv
the verb is present perfect;\footnote{There is a grammatical
temporal incoherence between \refegs{mare}{3}.iv and the preceding
discourse, as the former is in the present perfect, while the latter
is in the simple past. However, this temporal incoherence does not
affect resolution of pronoun reference, as we can change the tenses in
\refegs{mare}{1} and \refegs{mare}{2} to make the whole discourse
temporally coherent, and the same kind of pronominal reference occurs.
The coherent discourse is: \m{{\bf Maria}$_i$ vuole andare al mare.
{$\phi_i$} Ha telefonato a {\bf Giovanni}$_j$. $\phi_j$ Si e'
arrabbiato perche' $\phi_j$ stava dormendo.} which translates to
\m{{\bf Maria}$_i$ wants to go to the seaside. {\bf (She$_i$)} has
phoned to {\bf Giovanni}$_j$.  {\bf (He$_j$)} self is {become
angry$_{masc}$} because {\bf (he$_j$)} was sleeping.} The verb \m{volere},
\m{to want}, in the first clause cannot be used in the present perfect 
in this context.} the past participle agrees with the subject, and its
masculine morphology forces the null subject to refer to {\sf
Giovanni}, and not to {\sf Maria}.  This last alternation lends
support to my claim that it is the context up to and including the
verb that is taken into account when interpreting a zero: it is the
fact that the main verb is marked for masculine that allows the null
subject to refer to something different from Cb\refegs{mare}{2}.
\edes

\noindent

Further evidence for the importance of clues up to and including the
verb comes from clitics, more specifically, from clitics embedded in a
modal or control verb construction, as Exs.~\refegs{climb}{3}.i
through iii show.  The crucial features is that clitics may be
cliticized to the infinitival complement of the higher verb, 
as in \refegs{climb}{3}.i and  \refegs{climb}{3}.ii, or can
climb in front of the higher  verb, as in \refegs{climb}{3}.iii.

\begin{egs}{climb} 
\item \gll $\phi_k$ ho parlato con {\bf Maria}$_i$ ieri.\\
{\bf (I$_k$)} have talked with {\bf Maria}$_i$ yesterday.\\
\item \gll $\phi_i$ \'{E} arrabbiata$_{fem}$ con  {\bf Giorgio}$_j$:\\
{\bf (She$_i$)} is angry$_{fem}$ with {\bf Giorgio}$_j$:\\
\item 
\renewcommand{\theenumi}{\roman{enumi}}
\benum 
\item \gll  $\phi_i$ non vuole pi\'{u} parlar\/{\bf gli}$_j$.\\
{\bf (she$_i$)} not wants {any more} {to talk {\bf to him}$_j$.}\\
\item   \gll {\rm \#} $\phi_j$ non vuole pi\'{u} parlar\/{\bf le}$_i$.\\
\# {\bf (he$_j$)} not wants {any more} {to talk {\bf to her}$_i$.}\\
\item  \gll $\phi_j$ non {\bf le}$_i$ vuole pi\'{u} parlare.\\
{\bf (he$_j$)} not {\bf to-her}$_i$ wants {any more} {to talk}. \\
\eenum
\end{egs}

\bdes
\item[\refegs{climb}{3}.i] realizes a \centc, with the null subject referring to
Cp\refegs{climb}{2}~=~Cb\refegs{climb}{2}, namely {\sf Maria}. 
\item[\refegs{climb}{3}.ii] is incoherent. The preferred interpretation 
for the null subject is {\sf Maria}; however, when the clitic {\sf le}
is found at the end of the sentence, the hearer is forced to change
the interpretation of the null subject to {\sf Giorgio}. The effect is
similar to a syntactic ``garden path''.
\item[\refegs{climb}{3}.iii] is acceptable, as the clitic {\sf le},
that in \refegs{climb}{3}.ii is cliticized onto {\sf parlare}, climbs
in front of the modal verb {\sf vuole}: so the hearer is forced to
exclude {\sf Maria} as referent of the null subject. This happens {\em
early enough} so that no ``garden path'' effect is registered.
\edes
It is from the contrast between \refegs{mare}{3}.ii and
\refegs{mare}{3}.iv, and between the three \refegs{climb}{3}
variations that my claim about the importance of the context up to and
including the verbal complex stems.
Clearly, what exactly the
\m{context up to and including the verbal complex} amounts to is not
clear, as it includes agreement features, such as in
\refegs{mare}{3}.iv, but not the clitic which hasn't climbed, as in
\refegs{climb}{3}.i and \refegs{climb}{3}.ii: it apparently includes all the
verbal forms carrying tense and agreement features, which explains why
the past participle, marked for gender and number, is included, while
the infinitival complement of a modal or control verb is not.  The
observation that the context up to and including the verbal complex
affects the interpretation of the subject also makes sense from a
lexical semantics point of view, given that the lexical semantics of
the verb affects pronoun interpretation.

As we will see in Sec.~\ref{sec-clitic}, I found only few examples of
such configurations in the corpus: while they support the hypothesis
in \refegs{strategies}{2}, more data is required to come to definitive
conclusions.  Moreover, it is clear that psycholinguistic experiments
are needed to determine, among others: whether cases such as
\refegs{mare}{3}.iv, a \centss, require more time to process than
cases such as \refegs{mare}{3}.i, a \centc, even if both involve a
null subject; whether a \centss\/ overtly marked by a strong pronoun
such as in \refegs{mare}{3}.iii requires less processing time that one
encoded by a zero and supported by syntactic features, such as in
\refegs{mare}{3}.iv; whether indeed effects analogous to syntactic
garden paths occur in cases such as \refegs{climb}{3}.ii.
\label{est1}

\section{Corpus}

\label{corpus}
The corpus amounts to about 12,000 words (roughly 25 pages of
text). It is composed of excerpts from two books
\cite{giardino,penelope}, a letter \cite{mila}, a posting on the
Italian electronic bulletin board \cite{bboard}, a short story \cite{cicala}, and
three articles from two newspapers \cite{metz,dick,perot}.  The
excerpts are of different lengths, with the excerpts from the two
books being the longest, \cite{giardino} with 3,641
words\footnote{Most of my examples will come from \cite{giardino}, as
it is the largest source of pronominal expressions: from now on assume
that the source of an example is \cite{giardino}, unless otherwise
noted.} and \cite{penelope} with 1,918, and the posting on the Italian
bulletin board, with only 603 words, the shortest. 
Texts were chosen to cover a variety of contemporary written Italian
prose, from formal (newspaper articles about politics and literature),
to informal (posting on the Italian electronic bulletin board). 

The corpus I'm reporting about is a subset of the initial materials I
assembled.  In fact, I had to choose prose that describes situations
involving several animate referents, as strong pronouns can refer only
to animate referents.\footnote{On the contrary, null
subjects can refer to inanimate subjects as well and even be used for
discourse deixis, i.e. to refer to a whole preceding discourse segment
\cite{pennling89}.} Moreover,  I eliminated  texts that contain direct
speech, another thorny issue for centering and in general for theories
of discourse processing;  excerpts from the two books
don't contain dialogues. \cite{giardino} is a book written in the form
of a diary, which explains the presence of first person pronouns; the
diary format has the advantage that there are almost no dialogues,
which instead appear in the usual novel format.  The excerpts from
\cite{giardino} involve at least two people, and possibly at least two
are of the same sex; the chosen excerpts are discontinuous because,
first, the diary format is in itself discontinuous; second, as the
book revolves around the author's interest in gardening, many pages
discuss plants and flowers or describe the landscape, and they
obviously don't provide the required animate referents. The excerpt from
\cite{penelope} was chosen because the situation described involves
four people, two men and two women.

\subsection{Quantitative data}

Tables~\ref{numbers} through \ref{numbers4} provide the quantitative
data for each text.

\tref{numbers}  gives the total number of grammatical subjects, and 
out of these, the total number of animate subjects: I counted subjects
in both main and subordinate tensed clauses, but I excluded impersonal
constructions and relative clauses where the relative pronoun is the
subject.
\begin{table}[htpb]
\begin{center}
\begin{tabular}{||l||c||c||}\hline \hline
Text & Total & Animate \\ 
& subjects & subjects  \\  \hline 
\cite{giardino} & 241 & 229  \\ 
\cite{penelope} & 27 & 24  \\
\cite{mila} & 63 & 50 \\
\cite{bboard} & 45 & 32 \\ 
\cite{cicala} & 77 & 60  \\
\cite{metz} & 73 & 42  \\
\cite{dick} & 39 & 23 \\
\cite{perot} & 65 & 37  \\  \hline 
Total & 630 & 497  \\  \hline \hline
\end{tabular}
\end{center}
\protect \caption{Total and animate subjects}
\label{numbers}
\end{table}

\tref{numbers2} partitions animate subjects according to grammatical person.
I didn't distinguish between singular and plural pronouns, as no phenomenon
I will talk about seems to be affected by such distinction. 
About 90\% of referential expressions are singular, as there are 60 plural 
subjects out of 630 total subjects.

\begin{table}[htpb]
\begin{center}
\begin{tabular}{||l||c||c|c|c||}\hline \hline
Text & Total  & 1st  & 2nd & 3rd   \\  \hline 
\cite{giardino} & 229 &  73 & 0 & 156 \\ 
\cite{penelope} & 24 & 0 & 0 & 24 \\
\cite{mila} & 50 & 23 & 18 & 9 \\
\cite{bboard} & 32 & 9 & 1 & 22 \\
\cite{cicala} & 60 & 0 & 0 & 60 \\
\cite{metz} & 42 & 0 & 0 & 42 \\
\cite{dick} & 23 & 0 & 0 & 23  \\
\cite{perot} & 37 & 0 & 0 & 37  \\\hline 
Total & 497 & 105 & 19 & 373 \\\hline \hline
\end{tabular}
\end{center}
\protect \caption{First, second and third person animate subjects}
\label{numbers2}
\end{table}

\tref{numbers3} shows third person subjects partitioned into four
classes: full NPs ---  this category also covers NPs that
include a possessive adjective referring to an animate entity, which I will
discuss below; strong pronouns; null subjects; and  other anaphors,
such as {\sf uno}, \m{one$_{masc}$}, or {\sf tutte}, \m{all$_{fem}$}.

\begin{table}[htpb]
\begin{center}
\begin{tabular}{||l||c||c|ccc||}\hline \hline
Text & Total & Full NPs & Strong & Null & Other \\ \hline
\cite{giardino} & 156 &  45 &  23 & 81 & 7 \\ 
\cite{penelope} & 24 &  6 & 2 & 16 & 0 \\
\cite{mila} & 9 &  1 & 2 & 5 & 1 \\
\cite{bboard} & 22  & 7 & 0 & 11 & 4 \\
\cite{cicala} & 60 &  26 & 1 & 33 & 0 \\
\cite{metz} & 42 & 28 & 1 & 12 & 1 \\
\cite{dick} & 23 &  19 & 3 & 1 & 0 \\
\cite{perot} & 37 &  27 & 1 & 7 & 2 \\\hline
Total & 373 & 159 & 33 & 166 & 15 \\\hline \hline
\end{tabular}
\end{center}
\protect \caption{Distribution of 3rd person subjects}
\label{numbers3}
\end{table}

\begin{table}[htpb]
\begin{center}
\begin{tabular}{||l||c|c||}\hline \hline
Text & strong & null   \\  \hline 
\cite{giardino} &  23 & 36 \\ 
\cite{penelope} & 2 & 9 \\
\cite{mila} &  2 & 4 \\
\cite{bboard} & 0 & 7  \\
\cite{cicala} & 1 & 13 \\
\cite{metz} & 1 & 6 \\
\cite{dick} & 3 & 0 \\
\cite{perot} & 1& 5 \\ \hline
Total & 33 & 80 \\\hline \hline
\end{tabular}
\end{center}
\protect \caption{Strong pronouns and null subjects}
\label{numbers4}
\end{table}

Looking at \tref{numbers3}, it is apparent that the percentage of full
NPs versus pronouns is not constant through the eight texts.  The
percentages vary from 11\% in \cite{mila},\footnote{\cite{mila} is
probably a case in itself as it is a personal letter, and so it
employs many more first and second person pronouns than third person
ones --- see \tref{numbers2}.} to between 20\% and 30\% in
\cite{giardino}, \cite{penelope}, and \cite{bboard}, 21\%, 25\% and
27\% respectively. Then there is an increase for the last four texts,
from 43\% in \cite{cicala}, to 66\% in \cite{metz}, 72\% in
\cite{perot} and finally 82\% in \cite{dick}. Intuitively, it makes
sense that more formal prose employs longer and more elaborate
constructions.  

It is clear that a full analysis should include full NPs as well, as
about 60\% of the full NPs in \tref{numbers3} are used referentially:
for example, \cite[ch. 6]{turan95} discusses some intriguing results
regarding the referential usage of full NPs in subject position in
Turkish.  Turan notices that, in her Turkish corpus, \centrs's are
realized 99\% of the times by means of full NPs, and never by means of
a null subject --- this is consistent with the absence of \centrs's in
my pronominal data, see below.  For \centss's, the picture is more
complicated: Turan notices that the shift to the object of the
previous utterance is performed by means of a full NP if the object is
inanimate, of an overt pronoun if the object is animate. I have
started analyzing full NPs and their relation to centering: while I
won't discuss full NPs in this chapter, some preliminary results can
be found in \cite{coling96}.

Finally, \tref{numbers4} shows just the data of interest, namely 
strong pronouns and null subjects. Notice that the null subjects in
\tref{numbers4} amount to about half of those appearing in
\tref{numbers3}: to analyze centering transitions, I only considered
those null subjects whose antecedents are not determined by
contraindexing constraints \cite{lasnik76,chomsky81}. I also excluded
those
that appear in a conjoined main clause which is not the first
conjunct, and such that the null subject corefers with the subject of
the preceding conjunct:
\begin{eg}{free-main}
\item 
\gll {\bf Lui$_i$} non sembra mai demoralizzato, e $\phi_i$ va avanti ...\\
{\bf He$_i$} not appears ever frustrated, and $\phi_i$ carries on ... \\
\end{eg}
In this case I consider the null subject to be constrained as if by
contraindexing. Conjunctions do impose syntactic constraints that are
different from those derived from simply juxtaposing
clauses,\footnote{Contra \cite{kame93}.} as shown for example by the
fact that this is one of the rare contexts in which subject pronouns
are sometimes dropped even in English.

\section{Subject pronouns and centering transitions}

\label{cent-trans}

\subsection{Applying centering to real text}

When analyzing real text, one realizes that many issues are still
open. I will comment here on how deictics, possessives, and
subordinate clauses affect centering.

\paragraph{Deictics.} In texts such as
\cite{giardino}, \cite{mila} or \cite{bboard} there is an abundance of
first and second person pronouns, most of them singular (see
\tref{numbers}).  The problem is whether \m{situational deictics} such
as \m{I} and \m{you} are part of the Cf list or not; moreover, \m{I} in
\cite{giardino} often appears with verbs of thought, so that the
problem of how to deal with situational deictics compounds with the
problem of how to take subordinates into account.
\label{dm-page}
\noindent
Consider the following example, where, in the utterance preceding
\refegs{deictic-shift}{1}, Cb~=~Cp~=~{\sf pastore} (pastor$_{masc}$), 
and {\sf lui} in \refegs{deictic-shift}{2} refers to {\sf pastore}:
\begin{egs}{deictic-shift}
\item 
\gll {\bf Mi$_i$} e' venuto spesso di pensare che cosa terribile sarebbe\\
{{\bf To me$_i$}} is come often to think what thing terrible {would be}\\
\item
\gll se {\bf lui$_j$} si sentisse male nella sua bussola\\
if {\bf he$_j$} self$_j$ felt  bad {in the} his {small room}.\\
\end{egs}

The issue is whether \m{I} belongs to the Cf list of 
\refegs{deictic-shift}{1}, or of \refegs{deictic-shift}{1} and
\refegs{deictic-shift}{2} taken together, if a complement clause 
such as \refegs{deictic-shift}{2} is not an independent centering
unit. I follow \cite{lyn-workshop} in
assuming that deictics are always available as part of global focus,
and therefore are outside the purview of centering.

\paragraph{Possessives.}

As noted above, the full NP category in \tref{dist-total} includes NPs
that include a possessive adjective referring to an animate entity,
such as \m{i suoi sforzi} --- \m{his efforts}.  Possessives frequently
occur --- they constitute about one fifth of the full NPs that perform
centering transitions --- and provide another means of keeping the
center of attention. 

The problem is deciding how possessives affect Cb computation and the
order on the Cf list. An NP of type \m{possessive} in fact refers to
two entities, the possessor P$_{or}$ and the possessed P$_{ed}$: in
the following example, P$_{or}$~=~{\sf Irais$_i$} and P$_{ed}$~=~{\sf
husband$_k$} --- in the previous sentence, Cb={\sf Irais$_i$},
Cf=[{\sf Irais$_i$ $>$ English gentleman$_j$}]:
\begin{egs}{porta}
\item \gll {\bf Suo$_i$} {\bf marito$_k$} non ha pi\'{u} avuto pace,\\
{\bf Her$_i$} {\bf husband$_k$} not has {any longer} had peace,\\
\item \gll e ogni volta che {\bf $\phi_i$} deve uscire da una stanza ...\\
and every time that  {\bf (she$_i$)} {has to} leave from a room ...\\
\end{egs}
While Cb computation does not appear to be affected by a possessive,
that behaves like a pronoun, the Cf ranking needs to be modified.
P$_{ed}$ corresponds to the full NP, and thus its position in Cf is
determined by the NP's grammatical function; as regards P$_{or}$, my
working heuristics is to rank it as immediately preceding P$_{ed}$ if
P$_{ed}$ is inanimate, as immediately following P$_{ed}$ if P$_{ed}$
is animate.  Consider the following (contrived) discourse:\\
\begin{egs}{poss}
\item {\em I met {\bf Mary$_i$} yesterday. }
\item {\bf She$_i$} {\em was worried.}
\item  \renewcommand{\theenumi}{\roman{enumi}}
\begin{enumerate}
\item {\bf Her$_i$ husband$_j$} {\em was in the hospital.}
\item {\bf Her$_i$ car$_k$} {\em wasn't working.}
\end{enumerate}
\end{egs}
In both \refegs{poss}{3}.i and \refegs{poss}{3}.ii the Cb is {\tt
Mary$_i$\/}; as regards the Cf list, in \refegs{poss}{3}.i it is {\tt
[husband$_j$~(P$_{ed}$)~$>$ Mary$_i$~(P$_{or}$)]}, while in
\refegs{poss}{3}.ii it is {\tt
[Mary$_i$~(P$_{or}$)~$>$ car$_k$~(P$_{ed})$]}. Clearly this heuristics
needs to be rigorously tested.

\paragraph{Subordinates.}

Another important issue, that has not been extensively addressed yet
--- but see \cite{kame97}, \cite{suri93} --- is how to deal with complex
sentences that include coordinates and subordinates. The questions
that arise concern whether there are independent Cb's and Cf lists for
every clause; if not, how  the Cb of the complex sentence is computed,
and how  semantic entities appearing in different clauses are ordered
on the global Cf list.  

A simple example is the following discourse, for which I provide a  
literal, but not word by word, translation; 
for the utterance preceding \refegs{pigroni}{1}, we have
Cb(U$_{i-1}$)~=~{\sf vicina$_j$} (neighbor$_{fem}$),
Cf(U$_{i-1}$)~=~{\sf [vicina$_j$]}. 
\begin{egs}{pigroni}
\item   {\em  Prima che {\bf i} {\bf pigroni$_i$} siano seduti a 
tavola a far colazione,}\\
Before  {\bf the lazy ones$_i$}  sit down to have breakfast,
\normalsize
\item  {\em {\bf lei$_j$} e' via col {\bf suo$_j$} calessino alle altre cascine della tenuta.}\\
{\bf she$_j$} has left {with } {\bf her$_j$} buggy for the other farmhouses on
the property.
\normalsize
\end{egs}
The issue is whether the Cb and Cf list are updated after the whole
sentence, or whether a new Cb and Cf list are computed after
\refegs{pigroni}{1}: these new items would then be the input to a new
computation of Cb and Cf list after \refegs{pigroni}{2}.  It is my
impression that preposed adjuncts, such as \refegs{pigroni}{1}, do
affect centering transitions: the fact that an overt subject is used
after a preposed adjunct seems to support the fact that this is a
\cents\/ or a {\sc cent-est} --- see below --- and not a simple \centc\/
from the previous utterance that could be encoded with a null subject,
which on the contrary is not particularly felicitous here.  As a
working hypothesis, I've loosely adopted Kameyama's proposal 
\shortcite{kame93,kame97}, that sentences containing
conjuncts and tensed adjuncts are broken down into a linear sequence
of centering ``units'', while tenseless adjuncts don't generate
independent centering units\footnote{The situation for complements is
more complicated, and space prevents me from discussing it.}. Thus
\refegs{pigroni}{1} and \refegs{pigroni}{2} have each distinct Cb's and  Cf
lists.

\subsection{Centering Transitions}

Table~\ref{dist-total} shows the distribution of null and strong
pronouns with respect to centering transitions, while 
\tref{dist-gia} gives the distribution of transitions per text.
\begin{table}[htpb]
\small
\begin{center}
\begin{tabular}{||l||c||c|c|c||c||c||} \hline \hline
Type & Total &  {\sc continue} & {\sc 
retain} & {\sc shift} & {\sc Cent-est} & {\sc Other} \\ \hline \hline
zero & 80 & 56 & 4 & 6 & 12 & 2  \\ 
strong & 33 & 13 & 4 & 5 & 10 & 1 \\ \hline\hline
Total & 113 & 69 & 8 & 11 & 22 & 3 \\ \hline \hline
\end{tabular}
\end{center}
\protect \caption{Distribution of centering transitions}
\label{dist-total}
\end{table}
\normalsize

\begin{table}[htpb]
\small
\begin{center}
\begin{tabular}{||l||l||c||c|c|c||c||c||} \hline \hline
Text & Type & Total &  {\sc continue} & {\sc 
retain} & {\sc shift} & {\sc Cent-est} & {\sc Other} \\ \hline \hline
\cite{giardino} & null  & 36 & 20  & 4 & 5 & 6 & 1  \\ 
&strong& 23 & 7 & 3 & 3 & 9 & 1  \\ \hline
\cite{penelope} & null  & 9 & 7  & 0 & 0 & 2 & 0  \\ 
&strong& 2 & 0 & 0  & 1 & 1 & 0 \\ \hline
\cite{mila} & null  & 4 & 4  & 0 & 0 & 0 & 0  \\ 
&strong& 2 & 1 &  0 & 1 & 0 & 0 \\\hline
\cite{bboard} & null  & 7 & 6  & 0 & 0 & 1 & 0  \\ 
&strong & 0 & 0 &  0 & 0 & 0 & 0 \\ \hline
\cite{cicala} & null  & 13 & 11  & 0 & 0 & 2 & 0 \\ 
&strong& 1 & 1  & 0 & 0 & 0 & 0  \\ \hline
\cite{metz} & null  & 6 & 4  & 0 & 1 & 0 & 1  \\ 
&strong& 1 & 1 &  0 & 0 & 0 & 0  \\ \hline
\cite{dick} & null  & 0  & 0  & 0 & 0 & 0 & 0  \\ 
&strong& 3 & 2  & 1 & 0 & 0 & 0 \\ \hline
\cite{perot} & null  & 5 & 4   & 0 & 0 & 1 & 0 \\ 
&strong& 1 & 1 &  0 & 0 & 0 & 0 \\ \hline \hline
Total & & 113 & 69 &  8 & 11 & 22 & 3 \\ \hline \hline
\end{tabular}
\end{center}
\protect \caption{Distribution of centering transitions per text}
\label{dist-gia}
\end{table}

Tables~\ref{dist-total} and \ref{dist-gia}   require some explanation, as
they don't distinguish between {\sc smooth-} and  {\sc  rough-shift}, and
include new transitions such as {\sc cent-est}. 

First of all, 
I don't distinguish between {\sc smooth-} and {\sc rough-shift}, 
as {\sc rough-shift}'s involving pronouns can appear only 
in very specific conditions, that do not occur in my data. In fact, the
conditions for a \centrs\/ are: 
\benum
\item Cb(U$_i$)~$\neq$~Cb(U$_{i-1}$) and
\item Cb(U$_i$) $\neq$ Cp(U$_i$) \eenum Notice that given the Cf
ranking in \refeg{cf-ranking2}, the null or strong pronoun $p_i$ in
subject position will always be Cp(U$_i$).\footnote{Empathy effects
don't occur in my data when the subject is a pronoun: rather, they
arise when the subject is a full NP's pertaining to a character's
point of view, as in \m{Le sue convinzioni lo trascinano fuori
dalla casetta a tutte le ore --- His beliefs drag him out of his house
at all hours}.} Thus, for a {\sc rough-shift} to arise, $p_i$ must not
be Cb(U$_i$), otherwise a \centss\/ would occur.
For  $p_i$ not to be  Cb(U$_i$),  U$_i$ must have  at least another
pronoun (otherwise if $p_i$ is the only pronoun, it is Cb(U$_i$), and
being also Cp(U$_i$), condition 2 does not obtain).  A configuration in
which a \centrs\/ obtains in U$_{i}$ is, schematically --- both e$_2$
and e$_3$ are pronouns, e$_3$ corresponds to $p_i$:
\begin{center}
\begin{tabular}{lll}
U$_{i-1}:$ & Cb~=~e$_1$, & Cf~=~[e$_1$ $>$ e$_2$ $>$ e$_3$] \\
U$_{i}$: & Cb~=~e$_2$, & Cf~=~[e$_3$ $>$ e$_2$]
\end{tabular}
\end{center}
A constructed example where \refegs{rough}{2} and \refegs{rough}{3}
instantiate this configuration is:
\begin{egs}{rough}
\item 
\gll {\bf Giorgio$_i$} e' amico di {\bf Maria$_j$}.\\
{\bf Giorgio$_i$} is friend of {\bf Maria$_j$}.\\
\glt  \small {\tt Cb~=?; Cf~=~[Giorgio$_i$ $>$ Maria$_j$]}\\
\item
\gll $\phi_i$ {\bf l$_j$'} ha presentata$_{fem}$ a {\bf Giovanni$_k$}.\\
{\bf (He$_i$)} {\bf her$_j$} has introduced$_{fem}$ to {\bf Giovanni$_k$}.\\
\glt \small {\tt Cb~=~Giorgio$_i$; Cf~=~[Giorgio$_i$ $>$ Giovanni$_k$ $>$ Maria$_j$]}\\
\item
\gll  {\bf Lei$_j$} {\bf lo$_k$} trova antipatico.\\
{\bf She$_j$} {\bf him$_k$} finds unpleasant.\\
\glt \small {\tt Cb~=~Giovanni$_k$; Cf~=~[Maria$_j$ $>$ Giovanni$_k$]}\\
\end{egs}
However, there are no examples of this sort in my data.\footnote{Note that
\refegs{rough}{3} also has another interpretation, a \centr\/ with 
{\bf lo} referring to Giorgio$_i$ rather than to Giovanni$_k$: in this
case though, the \centrs\/ is  preferred to \centr. It is clear that
the semantics of the situation comes into play.}

Moving now to {\sc cent-est} and {\sc other}, also included in
Tables~\ref{dist-total} and \ref{dist-gia}, {\sc cent-est} --- for
{\sc center establishment} --- captures the fact that sometimes pronouns
(even the null subject!) can be used to refer to an entity in the
global focus, and not on the Cf list of the previous utterance. Also
\cite[p.216]{gjw95} notices this phenomenon:
\begin{quote}
The second case [of quasi violations of Rule 1] concerns the use of a
pronoun to realize an entity not in the C$_f$(U$_n$); such uses are
strongly constrained. The particular use that have been identified
involve instances where attention is shifted globally back to a
previously centered entity (e.g. \cite{grosz77}, \cite{reichman85}).
\end{quote}

However, not all occurrences of {\sc cent-est} represent a global
focus shift. For example, if one postulates that adjuncts constitute
centering units in themselves, it is possible that the shift in U$_i$
is to an entity that belonged to Cf(U$_{i-2}$), where U$_{i-1}$ is an
adjunct preposed to U$_i$: clearly such a shift seems to have a less
dramatic effect (in terms of inference load) than a shift to an entity
that had been mentioned much earlier in the discourse. 

I suspect that there may be some correlation between how global the
shift is and the usage of a specific form, in particular the usage of
a full NP. Moreover, when the null subject is used for {\sc cent-est},
the resulting discourse may be slightly incoherent. For example, in
\cite[p.  70-71]{giardino}, the author describes the pastor and his
wife. After discussing the virtues of both, the author devotes the
next 10 (complex) sentences only to the pastor; in fact, the 10th
sentence doesn't talk about either of them. When in the 11th sentence
the author uses a null subject to refer to both, the effect is
slightly incoherent.  Sentences 9 through 11 are reported here with a
literal, but not word by word, translation.

\begin{egs}{pw}
\item 
\m{{\bf Lui$_i$} non parla mai di {\bf queste cose$_j$}, ma come {$\phi_j$}
potrebbero rimanere nascoste?\\}
{\bf He$_i$} never talks about {\bf these things$_j$}, but how could {\bf
(they$_j$)} remain hidden?
\item \m{Qui {\bf tutti$_k$} sanno tutto prima che la giornata volga al termine, e 
quel che $\phi_l$ mettiamo in tavola \'{e} assai pi\'{u} d'interesse generale del pi\'{u}
sbalorditivo capovolgimento politico.}\\
Here {\bf everybody$_k$} knows everything before the day comes to an end, and what 
{\bf (we$_l$)} 
have for dinner  is of much more  general interest than the most
surprising political change.
\item \m{$\phi_m$ Hanno un cottage spazioso, carino, con un bel pezzetto 
di terreno attiguo al cimitero.}\\ 
{\bf (They$_m$)} have a roomy, nice
cottage with a sizable piece of land next to the cemetery.
\end{egs}

\noindent
Strong pronouns are sometimes used to establish a
new center by selecting a member of a set available on the Cf list of
the previous utterance, as in \refeg{set} and \refeg{set2}.
In the utterance preceding \refegs{set}{1}, Cb~=~Cp~=~\{{\sf pastore
\& sua moglie}\}, \{pastor \& his wife\}:

\begin{egs}{set}
\item \gll $\phi_i$ Sono entrambi di un'austera devozione.\\
{\bf (They$_i$)} are both of {an austere} devotion.\\
\item \gll {\bf Lui$_j$} lavora nella sua parrocchia con nobile dedizione, and ...\\
{\bf He$_j$} works {in the} his parish with noble dedication, and ...\\
\end{egs}

\noindent
Another such example:

\begin{egs}{set2}
\item \gll $\phi$ Avevamo ormai finito il t\'{e}\\
{\bf (We$_j$)} had already finished the tea\\
\item \gll e {\bf lei$_i$} era salita {di sopra} a cambiar\/{\bf si$_i$} quando ...\\
and {\bf she$_i$} was gone upstairs to {change {\bf herself$_i$}} when ...\\
\end{egs}
It is debatable whether, once a set is available on the Cf list, also
its members are: however, a null subject would be infelicitous both in
\refegs{set}{2} and in \refegs{set2}{2}, thus providing weak evidence
that the members of sets on the Cf list are not themselves available
on the Cf list. I consider such usages of strong pronouns as {\sc
cent-est}'s.

Finally, {\sc other} refers to configurations that I have left
unanalyzed for the time being: they are characterized by parallelism,
or by expressions that  build a set out of Cb(U$_{i-1}$) and
some other entity, such as \m{sia lui che sua moglie} --- \m{both him
and his wife}. It is not clear how to deal with these
constructions within the centering framework yet.

\section{Discussion}

\label{discuss}

The reader will recall that the reason I conducted my corpus analysis was 
to verify the strategies in \refeg{strategies}, repeated here 
for convenience:
\begin{egs}{strategies2}
\item Typically, a null subject signals  a \centc, and a strong pronoun
a \centr\/ or a \cents.
\item A null subject can be felicitously used in cases of \centr\/
or \cents\/ if $U_{i}$ provides syntactic features that force the null
subject to refer to a particular referent and not to Cb$(U_{i-1})$.
Moreover, it is the syntactic context up to and including the verbal
form(s) carrying tense and / or agreement that makes the reference
felicitous or not.
\end{egs}

\noindent
I will now detail the results.

\subsection{\centc\/ after \centr}

\label{ret-cont-sec}

The first part of \refegs{strategies2}{1} --- 
null subjects used for \centc\/ --- 
is strongly supported. Zeros are used 80\% of the times, and there is
a significant difference ($\chi^2$~=~9.204, p~$<$~0.01) between zeros
and strong pronouns used in \centc\/ and zeros and strong pronouns
used in all  other transitions taken together --- see the
following contingency table.\footnote{$\chi^2$ test
results are reported here more as a source of suggestive evidence than
as strong indicators, as the observations in the corpus, which come
from only 8 authors, are not totally independent.}
\begin{table}[htpb]
\small
\begin{center}
\begin{tabular}{l|c|c} 
 &  {\sc continue} & {\sc all others}   \\ \hline 
zero &  56 & 24   \\ 
strong & 13 & 20  \\ 
\end{tabular}
\protect \caption{{\sc continue} vs. all other transitions}
\label{cont-table1}
\end{center}
\end{table}
\normalsize
Thus, in its use of null subjects for \centc\/, Italian behaves in the
same way as languages as diverse as Japanese \cite{kame,lyn3,shima95}
and Turkish \cite{turan95}, (Turan, this volume).  In fact, the usage
of zeros for \centc\/ seems to be a robust cross-linguistic
phenomenon.

However, as the 20\% percentage of strong pronouns used for \centc\/ is not
negligible, I set out to investigate which factors may affect such a
choice.  I analyzed the \centc's in my corpus, and I did find that one
relevant factor is the transition preceding the \centc\/ in question.
Consider \tref{centcr}, that shows the different possible transitions in the
utterance U$_{i-1}$ preceding the utterance U$_{i}$ in which a
\centc\/ occurs. 
The configuration in which a \centc\/ is preceded by a \centr, which I
call \centcr, differs from the other two because of the constraint
Cp(U$_{i-1}$)~$\neq$~Cb(U$_{i-1}$) in the \centr: this in a sense predicts
that the center will shift. But if a \centr\/ is followed by a \centc,
as in a \centcr, such prediction is not
fulfilled. 
\begin{table}[htpb]
\small
\begin{center}
\begin{tabular}{||l||c|c|c||}\hline \hline
& {\sc continue} & {\sc retain} & {\sc shift } \\ 
U$_{i-1}$&Cb$_{i-1}$~=~Cb$_{i-2}$&Cb$_{i-1}$~=~Cb$_{i-2}$&Cb$_{i-1}$~$\neq$~Cb$_{i-2}$
\\ &Cp$_{i-1}$~=~Cb$_{i-1}$&{\bf Cp$_{i-1}$~$\neq$~Cb$_{i-1}$}&Cp$_{i-1}$~=~Cb$_{i-1}$\\
\hline
U$_{i}$&\multicolumn{3}{c||}{Cb$_{i}$~=~Cb$_{i-1}$}\\
&\multicolumn{3}{c||}{Cp$_{i}$~=~Cb$_{i}$}\\ \hline \hline
\end{tabular}
\protect \caption{Centering transitions preceding a \centc}
\label{centcr}
\end{center}
\end{table}
\normalsize

Before providing quantitative support for the distinct behavior of
\centcr, I will illustrate this configuration with Ex.~\refeg{ret-cont},
which provides   two examples of \centcr: the first,  in
\refegs{ret-cont}{3}, is realized with a strong pronoun; the second, in
\refegs{ret-cont}{5}, is realized with a null subject. In the utterance 
preceding \refegs{ret-cont}{1}, Cb~=~{\sf Irais} and Cf~=~{\sf
[Irais]} --- the translation is literal but not word by word.

\begin{egs}{ret-cont}
\item {\em $\Phi_i$ Incomincer\'{o} a ricondurre 
il {\bf suo$_j$ pensiero} sui {\bf suoi$_j$ doveri} chiedendo\/{\bf le$_j$} ogni giorno}\\
\small
(I$_i$) will start to bring  {\bf her$_j$ thoughts} back to {\bf her$_j$ duties} by  
asking {\bf her$_j$} every day \\
{\tt Cf: [Irais$_j$ $>$ Irais$_j$'s thoughts, Irais$_j$'s duties], 
Cb:Irais, continue}
\item \normalsize
{\em come sta {\bf suo$_j$ marito$_k$}}.\\
\small
how {\bf her$_j$ husband$_k$} is.\\
{\tt Cf: [husband$_k$ $>$ Irais$_j$], Cb:Irais$_j$, retain}
\normalsize
\item {\em Non \`{e} che {\bf lei$_j$} {\bf gli$_k$} voglia granch\'{e} bene,} \\
\small
It's not the case that {\bf she$_j$} cares much about {\bf him$_k$}\\
{\tt Cf: [Irais$_j$ $>$ husband$_k$], Cb:Irais$_j$, continue}
\normalsize
\item {\em perch\'{e} {\bf lui$_k$} non corre ad aprir\/{\bf le$_j$} la porta \\}
\small
because {\bf he$_k$} doesn't run to open the door {\bf for her$_j$}\\
{\tt Cf: [husband$_k$ $>$ Irais$_j$], Cb:Irais$_j$, retain}
\normalsize
\item {\em ogni volta che $\Phi_j$ si alza per lasciare la stanza;}\\
\small
whenever {\bf (she$_j$)} gets up to leave the room.\\
{\tt Cf: [Irais$_j$], Cb:Irais$_j$, continue}
\normalsize
\end{egs}

Moving now to  the quantitative analysis of  \centcr,  
\tref{centcr-dist} shows how \centcr's affect
the usage of null and strong pronouns --- {\sc cont-cont} and {\sc
shift-cont} respectively refer to a \centc\/ preceded by
another \centc\/ or by a \cents.
\begin{table}[htpb]
\small
\begin{center}
\begin{tabular}{||l||c||c|c||} \hline \hline
Type & Total & {\sc cont-cont+} & {\sc ret-cont}  \\
&&{\sc shift-cont} & \\ \hline \hline
zero & 56 & 51 & 5   \\ 
strong & 13 & 7 & 6  \\ \hline \hline
Total & 69 & 58 & 11  \\ \hline \hline
\end{tabular}
\protect \caption{Pronoun occurrences for  \centcr}
\label{centcr-dist}
\end{center}
\end{table}
\normalsize

Compared to strong pronouns, null subjects are used 87\% of the times
for {\sc cont-cont} and {\sc shift-cont} taken together and only 45\%
of the times for \centcr. Moreover, if  a zero is 
used in a \centc, that \centc\/ is ten times more
likely to be  a {\sc cont-cont} or {\sc shift-cont} than  a
\centcr; in contrast, if a strong pronoun is used in a \centc, that \centc\/ is as likely 
to be a {\sc cont-cont} or a {\sc shift-cont} as a
\centcr. These trends in usage are confirmed by  a strongly 
significant difference between zeros and strong pronouns used in {\sc
cont-cont} plus {\sc shift-cont}, and zeros and strong pronouns used
in \centcr\/ ($\chi^2$~=~10.910, p~$<$~0.001).
Moreover, there is a very strongly significant difference between zeros
and strong pronouns used in {\sc cont-cont} plus {\sc shift-cont}, 
and zeros and strong pronouns used in all other transitions, including
\centcr\/ ---  $\chi^2$~=~16.922, p~$<$~0.001, see Table~\ref{cont-table3}. 
\begin{table}[htpb]
\small
\begin{center}
\begin{tabular}{l|c|c} 
 &  {\sc cont-cont} + & {\sc ret-cont} +   \\ 
&  {\sc shift-cont} & {\sc all others} \\ \hline 
zero &  51 & 29   \\ 
strong & 7 & 26  \\ 
\end{tabular}
\protect \caption{{\sc cont-cont} + {\sc shift-cont} vs. all other transitions}
\label{cont-table3}
\end{center}
\end{table}
Consistently, there is
no significant difference between zeros and strong pronouns used in
\centcr\/ and  zeros and strong pronouns used in transitions
different from \centc\/ --- $\chi^2$~=~0.292, p~$<$~0.7, see
Table~\ref{cont-table4}.  This suggests that \centcr's pattern more
like transitions different from \centc\/ than like other \centc's; in fact,
all transitions other than \centc\/ in Table~\ref{dist-total} present
a rough half-half split between zeros and strong pronouns, as do
\centcr's in Table~\ref{centcr-dist}.
\begin{table}[htpb]
\small
\begin{center}
\begin{tabular}{l|c|c} 
& {\sc ret-cont }  &   {\sc all others} \\
& & (excluding {\sc continue}) \\ \hline 
zero &  5 & 24   \\ 
strong & 6 & 20  \\ 
\end{tabular}
\protect \caption{{\sc ret-cont}  vs.  transitions different from {\sc continue}}
\label{cont-table4}
\end{center}
\end{table}
\normalsize

My results on different pronominal distributions for {\sc cont-cont}
and {\sc shift-cont} on the one hand, and {\sc ret-cont} on the other,
seem to be yet another source of evidence for the hypothesis that a
\centr\/ signals an upcoming shift: namely, not fulfilling the
prediction given by the Cp seems to require the explicit signal
provided by a strong pronoun. \footnote{The only researcher I know of
who argues against the prediction associated with a \centr\/ is
Linson: in \shortcite{linson93}, he presents evidence based on a
corpus study, in which a \centr\/ is followed by a \centc\/ 50\% of the
times, and by a \cents\/ only 15\% of the times. However, Turan notices
(p.c.)  that Linson used the Cf ranking in \refeg{cf-ranking}, and
that if \refeg{cf-ranking} is amended as in \refeg{cf-ranking2},
i.e. taking ``empathy'' into account, Linson's results may be
different, in that certain \centr's may in fact be \centc's.}

Also \cite{turan95} independently noticed the existence of {\sc
ret-cont}'s, and her results are compatible with mine: she found that
in {\sc ret-cont}'s, zeros decrease from 97\% to 68\% while strong
pronouns increase from 1\% to 11\% with respect to their percentages
of use for {\sc cont-cont} and {\sc shift-cont}.

\paragraph{\centr, \cents\/ and {\sc cent-est}.} As far as 
\centr's and \cents's
are concerned, the numbers are too small to draw any definitive
conclusion.  The tentative one is as follows: the examples I found do
seem to support \refegs{strategies2}{2}, as I will discuss below;
namely,  the null subject can be used in cases of \centr's and
\cents's if there are enough ``early'' clues that force the null
subject to refer to a particular referent. However, the numbers in
themselves do not identify any preferred usage for strong pronouns for
these transitions, contrary to what claimed by
\refegs{strategies2}{1}.

{\sc cent-est}'s  pattern like \centr\/ and  \cents\/
(and \centcr!), in that zeros and strong pronouns appear to be 
evenly distributed; moreover,   there is a
significant difference between zeros and strong pronouns used for 
{\sc cent-est}'s and zeros and strong pronouns used for {\sc cont-cont} 
plus {\sc shift-cont} ($\chi^2$~=~10.624, p~$<$~0.01).

A topic for future work is to investigate which factors, if any,
affect the choice between null and strong pronouns in these
configurations, especially because null subjects used for
\centss\/ or for {\sc cent-est} sometimes result in a slightly less
coherent discourse --- see \refegs{pw}{3}.

\subsection{Verb agreement, clitics, and null / strong pronouns}
\label{sec-clitic}

The second part of my claim, \refegs{strategies2}{2}, namely, that a
null subject can be used if U$_{i}$ provides syntactic clues that
force the null subject not to refer to Cb(U$_{i-1}$), is indeed borne
out --- however, given the small number of occurrences of null
subjects encoding these transitions (four \centr's and six {\cents}'s)
this conclusion can just be tentative. The most frequent clue is
agreement in gender and / or number; in some examples, clitics are
useful for disambiguation as well, but I found no example of clitic
climbing as discussed with respect to Ex.~\refeg{climb}.

However, I hoped to be able to verify a stronger claim, that whenever
such clues are available a null subject is used. But the data only
partly support this stronger claim. In fact, of the 9 instances of
strong pronouns realizing a \centr\/ or \cents, 4 do have clues that
should make a null subject possible.  Two of the four examples, both
\centr's, are :
\begin{egs}{strong-clue1}
\item 
\gll {\bf Io$_i$} faccio visita {\bf a lei$_j$} una volta all'anno,\\
{\bf I$_i$} pay visit {\bf to her$_j$} one time {per year},\\
\item
\gll e {\bf lei$_j$} {\bf mi$_i$} ricambia la visita quindici giorni dopo.\\
and {\bf she$_j$} {\bf to me$_i$} returns the visit fifteen days later.\\
\end{egs}
\begin{egs}{strong-clue2}
\item 
\gll $\phi_i$ \'{E} pronta a difender\/{\bf lo$_j$} in ogni occasione contro di {\bf noi$_k$}.\\
{\bf (She$_i$)} is ready to {defend {\bf him$_j$}} in every occasion against of {\bf us$_k$}.\\
\item \gll {\bf Lui$_j$} non {\bf le$_i$} parla mai.\\
{\bf He$_j$} not {\bf to her$_i$} talks ever.\\
\end{egs}
Both \refegs{strong-clue1}{2} and \refegs{strong-clue2}{2}
have a clitic available  before the main verb, analogously to
\refegs{climb}{3}.iii: thus, substituting  the strong pronoun with a zero
should result in a coherent discourse, but this is not the case. 
\cite{turan95} notices that in Turkish the rule that prescribes
using zeros in a \centc\/ is overridden if the pronominal expression
has to carry additional pragmatic information, such as phonetic
prominence or a listing reading. In the case of
\refegs{strong-clue1}{2}, clearly parallelism comes into play. In
\refegs{strong-clue2}{2}, there is indeed a contrast between {\bf lei}
(a female guest) defending {\bf lui} (the author's husband), and {\bf lui}
trying to ignore {\bf lei} as much as possible.

In contrast to \refegs{strong-clue1}{2} and \refegs{strong-clue2}{2},
I would like to mention two examples, again both  \centr's, that
pattern like Ex.~\refegs{climb}{3}.ii, namely, where the clitic is
found ``too late'' to allow the usage of a null subject.

The first example can be found in
\refegs{ret-cont}{4} above. The clitic {\bf le} (for her) is
cliticized to the infinitive \m{aprire} (to open), which is an adjunct
to the main verb \m{corre} (runs).
The second example is \refegs{nat-climb1}{4}: the clitic {\bf le} (to
her) doesn't climb in front of the modal
\m{vuole} (wants), but is  cliticized to the lower verb \m{correr(e)}
(to run).
\begin{egs}{nat-climb1}
\item \gll Ma {\bf lui$_k$} doveva sposare {\bf la} {\bf  cuoca$_{j,fem}$},\\
But {\bf he$_k$} {wanted to} marry {\bf the} {\bf cook$_{j,fem}$},\\
\item \gll e {\bf la} {\bf cuoca$_j$} ha visto un fantasma \\
and {\bf the} {\bf cook$_j$} has seen a ghost \\
\item \gll e $\phi_j$ {\bf se$_j$} n'\'{e} andata$_{fem}$ su due piedi,\\
and $\phi_j$ {\bf self$_j$} is gone on two feet,\\
\item  \gll e {\bf lui$_k$} vuole correr\/{\bf le$_j$} dietro ...\\
and {\bf he$_k$} wants {to run {\bf to her$_j$}} after ...\\
\end{egs}

\section{Conclusions}

\label{conclude}

The work presented in this chapter aims at explaining the different
usages of Italian pronominal subjects in terms of centering
transitions. The current research follows up on and extends
\cite{coling90}. My goal was to test the claims made in my earlier
work and based on constructed examples against naturally occurring
data. Not surprisingly, conducting a corpus analysis was useful not
just to verify those hypotheses but also to extend the analysis in a
variety of ways. The hypothesized strong preference for null subjects
in the case of \centc\/ is verified. Furthermore, taking the
transition preceding a \centc\/ into account provides an elegant
explanation for about half  of the strong pronouns used in
\centc's: a \centc\/ preceded by a \centr\/ behaves differently from one
preceded by a \centc\/ or by a \cents.

The results regarding the usage of strong pronouns for \centr\/ and
\cents\/ are mixed: in fact, the numbers don't indicate any preference
for one pronominal form over the other. Somewhat to my surprise, I
found that what is supported, at least tentatively given the small
numbers, is the second part of my claim, \refegs{strategies2}{2}: a
null subject can be used for \centr\/ or \cents\/ if the context up to
and including the verbal forms marked for tense and agreement provides
``early enough'' clues that prevent pronominal interpretation garden
paths.

It is clear that it is necessary to refine the analysis first of all
by collecting more instances of \centr's and \cents's, and of \centc's
occurring after \centr. Moreover, other pragmatic factors, such as
parallelism and contrast, should be examined, in order to understand
how they affect the choice of referring expressions.

I think a fruitful direction in which to move is to study the
functions of referential full NP's in terms of centering transitions. Some
preliminary results, available in \cite{coling96}, show that the
percentage of \centc's realized by means of full NP's is not
negligible at all, as it amounts to 16\%; and that full NP's account
for the majority of {\sc cent-est}: the preference for full NP's over
other referring expressions for {\sc cent-est} is statistically
significant. If {\sc cent-est}'s do correspond, at least in part, to
shifts in global focus, as mentioned in Sec.~\ref{cent-trans}, an
issue to tease apart concerns the conditions under which  full NP's,
strong pronouns and zeros are used. In general, by
including full NP's in the analysis, a more complete account of the
choices of referring expressions will be possible.

\subsection*{Acknowledgements}

This research was carried out while the author was with the
Computational Linguistics program, Carnegie Mellon University,
Pittsburgh, PA, USA. I wish to thank Susan Brennan, \"{U}mit Turan and Marilyn
Walker for their insightful comments on earlier versions of this
chapter.

\end{document}